%% file: CameraReady.tex
\documentclass[letterpaper]{article} 
\usepackage{aaai24}  
\usepackage{times}  
\usepackage{helvet}  
\usepackage{courier}  
\usepackage[hyphens]{url}  
\usepackage{graphicx} 
\usepackage{natbib}  
\usepackage{caption} 
\usepackage{url}

\usepackage{pdfpages}

\usepackage{algorithm}
\usepackage{algorithmic}
\usepackage{amsmath}
\usepackage{pdfpages}
\usepackage{newfloat}
\usepackage{listings}
\DeclareCaptionStyle{ruled}{labelfont=normalfont,labelsep=colon,strut=off} 
\floatstyle{ruled}
\newfloat{listing}{tb}{lst}{}
\floatname{listing}{Listing}

\title{}

\title{Reconstructing Persistent Worlds from Narratives for Narrative-Grounded Interactive Experiences}
\author {
    Yi-Chun Chen
}
\affiliations{
    National Cheng Kung University\\
    Tainan City 701, Taiwan\\
    rimi.chen@gs.ncku.edu.tw, ychen74@alumni.ncsu.edu
}

\usepackage{bibentry}

\begin{document}
\nocopyright
\maketitle

\begin{abstract}
Designing narrative-grounded interactive experiences remains labor-intensive because interactive content must align with the underlying world implied by the narrative. Existing approaches formulate problems such as narrative planning, scene generation, and gameplay generation, each constructing computational representations tailored to specific downstream tasks rather than explicitly reconstructing and maintaining the persistent world that grounds them.

We investigate reconstructing explicit persistent worlds from narrative descriptions as the central computational objective for narrative-grounded interactive realization. Rather than treating the world as an implicit by-product of downstream generation, our approach reconstructs and maintains persistent entities, locations, semantic relationships, and evolving world states while inferring only the contextual information required to support coherent interactive experiences.

To investigate this perspective, we develop a reference prototype that reconstructs structured persistent world representations from narrative descriptions and subsequently instantiates playable tile-based environments. Through three representative case studies spanning a procedural scenario, an original fantasy narrative, and an adapted public-domain story, we demonstrate the feasibility of reconstructing persistent worlds and show how a shared world representation supports coherent gameplay while remaining grounded in the source narrative.

By explicitly reconstructing persistent worlds prior to interactive realization, this work bridges computational narrative understanding and interactive content generation, providing a semantic foundation for AI-assisted game authoring, mixed-initiative design, educational simulations, and narrative-grounded interactive experiences.
\end{abstract}

\section{Introduction}

\input{1_Introduction}

\section{Persistent Worlds for Narrative-Grounded Interactive Experiences}
\label{sec:problem_formulation}

\input{2_Problem}

\section{Related Work}

\input{3_RelatedWork}

\section{Reference Framework}
\label{sec:framework}

\input{4_Framework}

\section{Reference Prototype}
\label{sec:implementation}
\input{5_Prototype}

\section{Prototype Validation}

\input{6_Evaluation}

\input{7_Discussion}

\section{Conclusion}

\input{8_Conclusion}
\bibliography{aaai24}

\newpage
\appendix
\includepdf[pages=-]{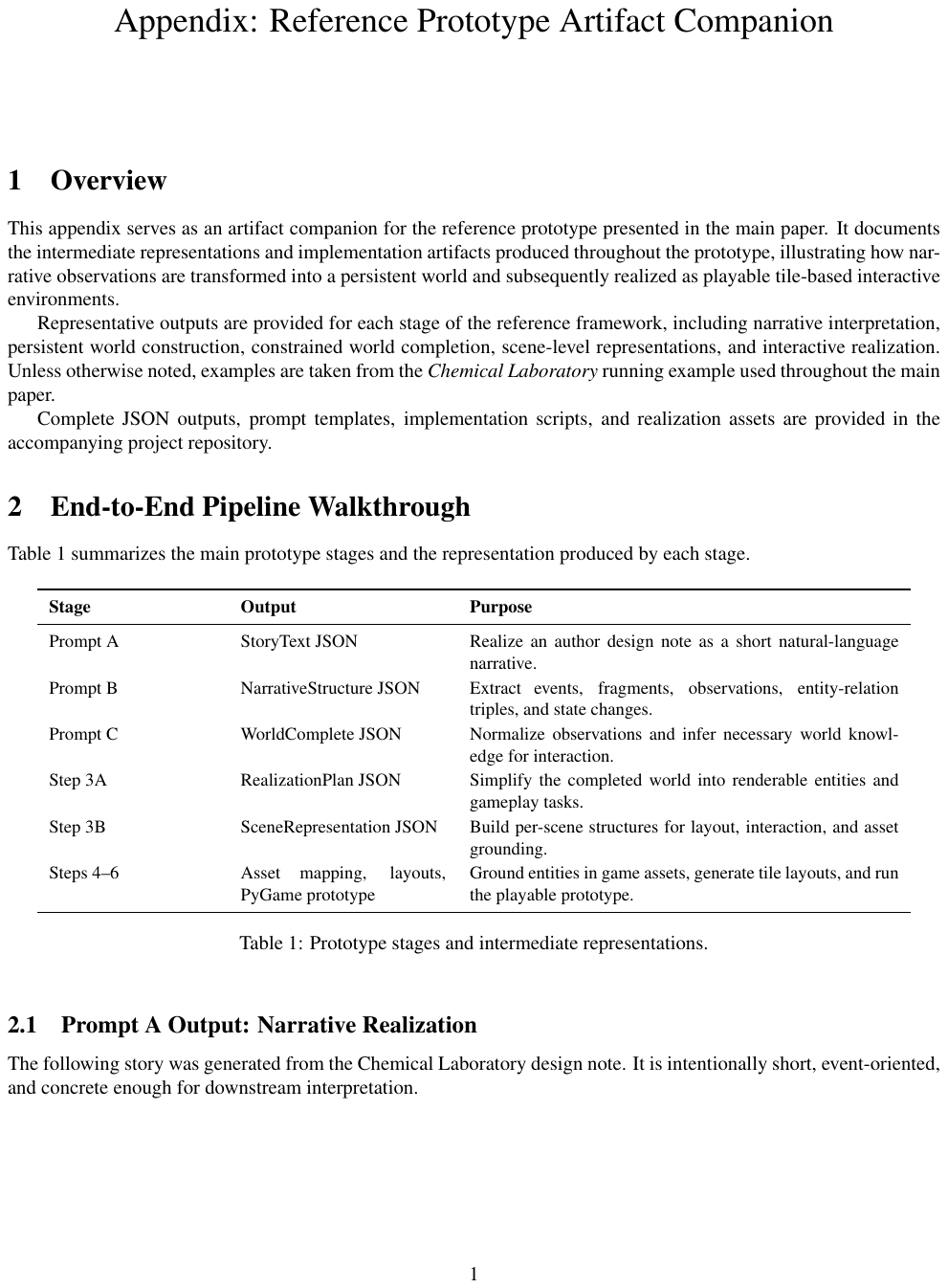} 

\end{document}

%% file: 1_Introduction.tex
Narratives provide a natural foundation for interactive experiences, including narrative-driven games, interactive storytelling, educational simulations, training environments, and mixed-initiative authoring systems~\cite{cardona2020gfi,horswill2014guest}. Rather than specifying every detail of an environment, narratives communicate an intended world through characters, events, locations, and their evolution over time. Designers subsequently construct environments, gameplay, and visual assets that realize this world while maintaining coherence with the source narrative~\cite{cardona2023aligning}. As AI increasingly assists game development, interactive content must remain grounded in the narrative world rather than merely being individually plausible.

Recent advances have introduced computational problems such as narrative planning~\cite{ware2021sabre}, scene generation from narrative descriptions~\cite{chen2025narrative}, gameplay generation, and end-to-end game generation~\cite{zhou2025story2game}. These approaches construct computational representations tailored to specific downstream objectives. However, narratives intentionally omit much of the contextual information required for interactive experiences, including persistent entities, spatial organization, and evolving world states. Consequently, downstream systems must reconstruct and maintain aspects of the underlying world before coherent interactive experiences can be realized.

\begin{figure*}[t]
\centering
\includegraphics[width=0.65\textwidth]{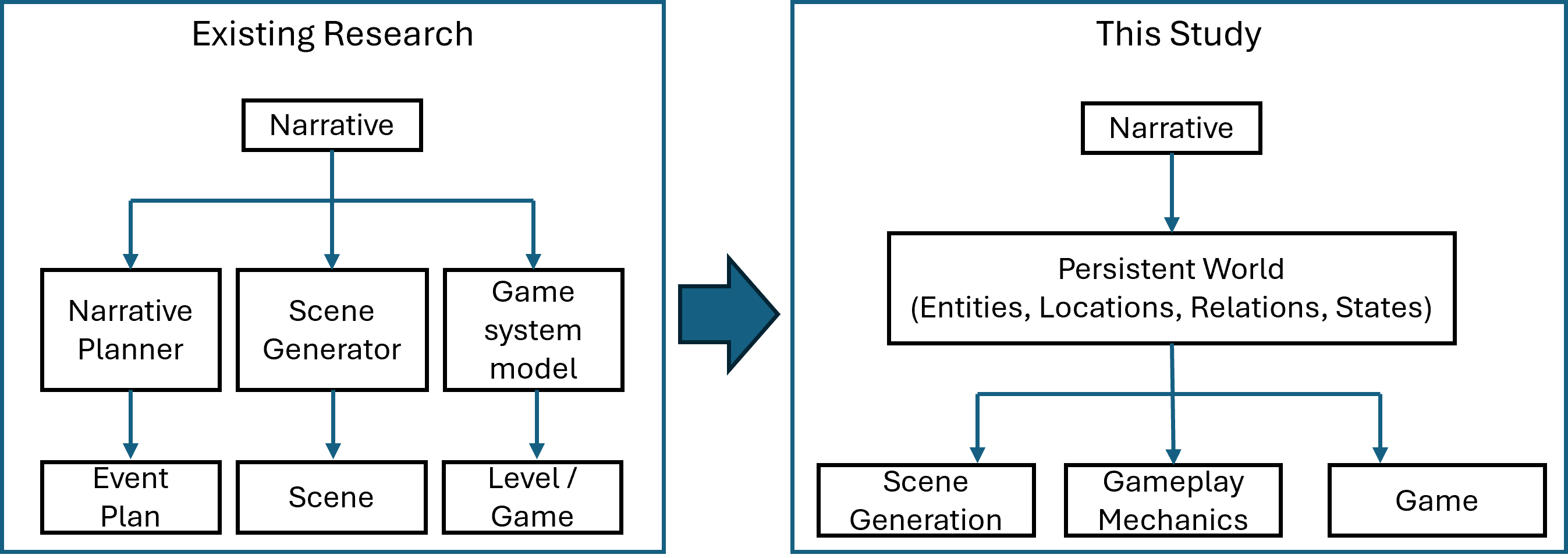}
\caption{Persistent world as a computational foundation for narrative-grounded interactive realization. Existing methods construct task-specific representations for downstream generation, whereas our framework reconstructs an explicit persistent world that supports multiple interactive realizations while preserving narrative consistency.}
\label{fig:teaser}
\end{figure*}

Although existing approaches necessarily reconstruct some representation of the underlying world, these representations are typically created to support a particular downstream objective, such as planning, scene generation, or gameplay realization. As a result, they primarily function as intermediate artifacts rather than reusable computational objects shared across subsequent computations. We instead argue that the persistent world implied by a narrative should be reconstructed explicitly before interactive realization. Unlike individual scenes or gameplay sequences, the persistent world maintains entities, locations, semantic relationships, and evolving world states across narrative events and player interactions. Figure~\ref{fig:teaser} illustrates this distinction between existing narrative-conditioned generation pipelines and our reconstruction-centered framework, in which an explicit persistent world serves as the computational foundation for interactive realization.

Motivated by this perspective, we investigate reconstructing explicit persistent worlds from narrative descriptions as the central computational objective for narrative-grounded interactive realization. We interpret narratives as partial observations of an underlying world rather than complete specifications of interactive content. The reconstructed world maintains persistent entities, locations, semantic relationships, and evolving world states while inferring only the contextual information required for coherent interactive realization. This reconstructed world subsequently serves as a reusable computational foundation from which multiple narrative-grounded interactive experiences can be instantiated.

The contributions of this paper are as follows:

\begin{itemize}

\item We investigate reconstructing persistent worlds from narrative descriptions as the central computational objective for supporting narrative-grounded interactive experiences.

\item We develop a reference prototype that reconstructs persistent worlds from narrative descriptions and instantiates playable tile-based environments.

\item Through representative case studies, we demonstrate the feasibility of persistent world reconstruction and show that a shared persistent world provides a reusable computational foundation for coherent narrative-grounded interactive realization.

\end{itemize}

%
%
%

%% file: 2_Problem.tex

This section formulates the need for reconstructing an explicit persistent world from narrative descriptions to support narrative-grounded interactive experiences. Rather than directly transforming narratives into scenes, levels, or complete games, we treat the \emph{persistent world} as the primary computational object between narrative understanding and interactive realization. We first define the scope of narratives considered in this work, explain why interactive realization requires a persistent world, define the role of that world, characterize narratives as partial observations, formulate persistent world reconstruction from narrative descriptions, and clarify how the reconstructed world supports downstream interactive realization.

\subsection{Narrative Scope}
\label{subsec:narrative_scope}

This work does not seek to provide a general linguistic or narratological definition of narrative. Instead, it defines the class of narratives addressed by the proposed formulation. We consider narratives that describe the evolution of an underlying world through a sequence of temporally related observations or events. Although each observation reveals only part of the world, the narrative collectively provides information about entities, relationships, locations, and states that persist or change over time.

Inputs that do not describe temporal world evolution, such as image captions or isolated scene descriptions, fall outside the scope of this formulation. Such inputs may support the realization of an individual scene, but they do not provide a sequence of observations from which persistence and change across events can be reconstructed. Table~\ref{tab:narrative_scope} summarizes the narrative types considered in this work.

\begin{table}[t]
\centering

\begin{tabular}{p{0.44\linewidth}|p{0.48\linewidth}}
\textbf{Included} & \textbf{Outside Scope} \\
\hline
Fictional stories & Image captions \\
Narrative quests & Single scene descriptions \\
Educational simulations & Standalone prompts \\
Procedural scenarios & Isolated observations \\
Historical narratives & Single moment descriptions \\
Role-playing game scripts & Standalone object descriptions \\
\end{tabular}

\caption{Examples of narrative inputs considered in this work and inputs outside its scope.}
\label{tab:narrative_scope}
\end{table}

\subsection{Why Interactive Realization Requires a Persistent World}
\label{subsec:need_persistent_world}

Narratives communicate enough information for readers to follow characters, events, and changes in the narrated situation, but they rarely specify everything required to construct an interactive experience. Interactive realization additionally requires decisions about which entities exist, where they are located, how locations are connected, which object states persist, what interactions are possible, and how actions affect later situations.

These requirements cannot be addressed independently for every scene or gameplay sequence without risking incompatible assumptions. An entity introduced in one event may need to remain available later; an object collected in one location may need to remain in the player's inventory after a transition; and an opened door or altered object state may need to persist across subsequent interactions. If each scene or downstream process reconstructs this information separately, individually plausible outputs may fail to describe one coherent world.

A shared persistent world provides the computational basis for coordinating these requirements. It preserves information communicated earlier, integrates later observations and interaction updates, and supplies a common reference for constructing scenes, determining possible actions, and maintaining continuity across an interactive experience. The need for a persistent world therefore arises not merely from a desire to store reusable information, but from the requirement that multiple narrative events and interactive realizations remain grounded in the same evolving world.

\subsection{Persistent World}
\label{subsec:persistent_world}

Narratives rarely describe complete worlds. Instead, they reveal observations that readers naturally interpret as occurring within a shared persistent context. Characters, locations, objects, and events introduced throughout a narrative are understood as belonging to the same underlying world, even though many aspects of that world remain implicit.

We define a \emph{persistent world} as the computational account of the narrative world that persists across narrative observations and subsequent interactions. Rather than corresponding to a particular scene or gameplay sequence, the persistent world maintains entities, locations, semantic relationships, and evolving world states that provide a common reference for interpreting observations and supporting interactive realization.

Here, the term \emph{world} does not imply a complete specification of every fact about a fictional universe. Instead, it refers only to the information required to maintain a coherent account of the entities, relationships, locations, and states relevant to the narrative and its interactive realization. The scope of the reconstructed world is therefore determined by the narrative observations together with the contextual information required for coherent interaction.

\begin{figure}[t]
\centering
\includegraphics[width=0.75\columnwidth]{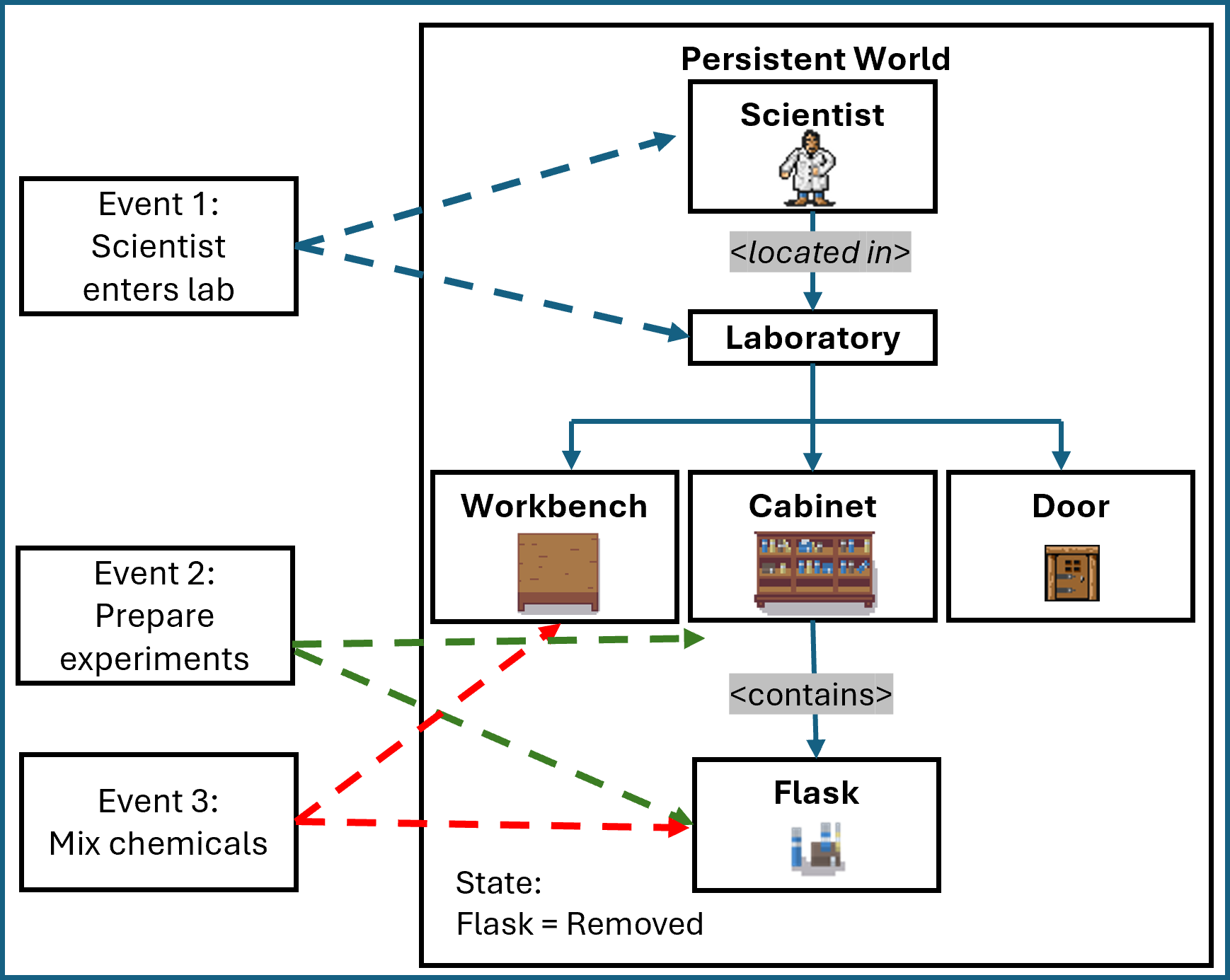}
\caption{Conceptual illustration of a persistent world. Narrative observations reveal only part of the underlying world, while the reconstructed persistent world maintains entities, locations, semantic relationships, and evolving world states that provide a shared foundation for multiple interactive realizations.}
\label{fig:persistent_world}
\end{figure}

This definition specifies the computational role of the persistent world rather than any particular implementation. A persistent world may be represented using knowledge graphs, symbolic world models, scene graphs, relational databases, or other structured representations. The proposed formulation is therefore independent of any particular representation. Unlike scenes, game levels, or gameplay sequences, which each realize only part of the underlying world, the persistent world serves as a reusable computational foundation from which multiple interactive realizations can be consistently instantiated. Figure~\ref{fig:persistent_world} illustrates this distinction.

\subsection{Narrative Observations}
\label{subsec:narrative_observations}

The proposed formulation interprets narratives as partial observations of a persistent world rather than complete specifications of that world. Authors communicate information relevant to the intended discourse while leaving much of the surrounding context implicit. Human readers combine explicit observations with contextual reasoning and prior knowledge to construct a coherent account of the characters, objects, locations, and states involved. In this paper, we use the following short procedural narrative as a running example.

\begin{center}
\fbox{
\begin{minipage}{0.92\linewidth}

A scientist entered a laboratory to conduct an experiment. After preparing the required equipment and materials, the scientist mixed the chemicals to complete the experiment.

\end{minipage}
}
\end{center}

The narrative explicitly describes the major activities and participating entities but omits much of the contextual information required to instantiate an interactive environment, including the laboratory layout, object placement, navigable space, supporting environmental structures, and the persistent states of objects as the experiment progresses. Nevertheless, readers can infer enough of this context to understand the narrated events.

The proposed formulation adopts the same perspective computationally. Rather than treating the narrative as a complete specification of an interactive environment, it treats the narrative as a sequence of observations from which a persistent world can be reconstructed. Explicit narrative observations provide the evidence for reconstruction. Constrained inference supplements only the contextual information required for coherent interaction, while previously established entities, relationships, and states are maintained unless later observations or interactions modify them. The running example is revisited throughout the remainder of the paper to illustrate these operations.

\subsection{Persistent World Reconstruction}
\label{subsec:persistent_world_reconstruction}

Based on the preceding definitions, we formulate \emph{persistent world reconstruction from narratives} as the computational task of constructing and maintaining an explicit persistent world from narrative observations. Let

\[
N=\{o_1,o_2,\ldots,o_n\}
\]

denote a narrative consisting of a sequence of temporally ordered observations. The objective is to reconstruct a persistent world

\[
W=(E,L,R,\Sigma),
\]

where \(E\) denotes persistent entities, \(L\) represents spatial organization, \(R\) captures semantic relationships, and \(\Sigma\) represents the evolving world state.

Formally,

\[
f:N\rightarrow W,
\]

where the reconstructed world satisfies the following properties.

\begin{itemize}

\item \textbf{Narrative fidelity.} Preserve information explicitly communicated by the narrative.

\item \textbf{Persistence and continuity.} Maintain entities, locations, relationships, and states across observations and subsequent interactions unless explicitly modified.

\item \textbf{Constrained reconstruction.} Reconstruct only the contextual information required to support coherent interactive realization.

\item \textbf{World consistency.} Maintain one coherent account of entities, locations, relationships, and states as narrative observations and interaction updates accumulate.

\item \textbf{Operational foundation.} Provide the maintained world required to support downstream interactive realization.

\end{itemize}

The formulation specifies the computational objective rather than prescribing any particular representation, reconstruction strategy, or implementation. Knowledge graphs, symbolic world models, relational databases, scene graphs, and hybrid representations are all compatible provided they satisfy the properties above. Persistent world reconstruction is therefore distinct from directly generating scenes, levels, or complete games. Those downstream realizations operate on the maintained persistent world rather than replacing it.





\subsection{Interactive Realization}
\label{subsec:interactive_realization}

Once a persistent world has been reconstructed, it serves as the computational interface between narrative understanding and interactive realization. It provides the entities, locations, relationships, evolving states, and interaction-relevant context from which individual scenes and gameplay situations can be instantiated.

The same persistent world may support multiple downstream realizations, including explorable game environments, educational simulations, interactive storytelling experiences, and mixed-initiative authoring systems. These realizations may differ in layout, presentation, mechanics, or interaction design while remaining grounded in the same narrative observations and maintained world.

Interactive realization may subsequently update the persistent world. Player actions can modify entity locations, inventory contents, object states, or semantic relationships, and these updates become part of the maintained world available to later scenes and interactions. The persistent world, therefore, connects narrative interpretation with continued interaction by serving as the long-lived computational state shared across both processes.

%% file: 3_RelatedWork.tex
This work builds upon research in computational narrative, interactive storytelling, narrative-conditioned content generation, procedural content generation, and mixed-initiative game design. Rather than organizing related work by application domain, we examine the computational representations constructed by existing approaches and the roles those representations play in supporting narrative-grounded interactive experiences. Our focus is not simply on whether prior work represents narrative or world information, but on whether it reconstructs and maintains an explicit, persistent world as the shared computational object that links narrative interpretation, interactive realization, and subsequent interaction.

\subsection{Narrative-Grounded Interactive Experiences}

Narratives underpin many interactive experiences by defining characters, events, locations, goals, and their evolution over time. Prior work has studied maintaining coherence between story progression and gameplay~\cite{cardona2023aligning}, formal narrative reasoning~\cite{cardona2020gfi,horswill2014guest}, experience management under player agency~\cite{ware2022multiagent}, and consistency across game artifacts generated with large language models~\cite{gallotta2024consistent}. These approaches establish narrative coherence as a central objective of AI-assisted interactive experiences.

Maintaining coherence, however, requires more than ensuring that individual scenes, actions, or artifacts are locally plausible. Information introduced in one narrative event may constrain later environments, interactions, and gameplay states. Characters and objects must retain their identities, locations and relationships must remain compatible, and changes produced by narrative events or player actions must persist across subsequent realizations. This work focuses on the explicit computational object needed to maintain that continuity: a persistent world shared across narrative interpretation and interactive realization.




\subsection{Computational Narrative and Story Understanding}

Computational narrative develops representations that support narrative understanding beyond surface text~\cite{montfort2023computational,gervas2024challenges,castricato2021towards}. Recent work models storyworlds through world-state transitions~\cite{gongora2026world}, represents narrative chronology~\cite{gervas2024representing}, and performs commonsense reasoning over event preconditions and effects~\cite{xie2025making}. Structured representations further support downstream applications, including hierarchical visual narrative understanding~\cite{chen2025hierarchical} and collaborative visual narrative generation~\cite{chen2024collaborative}.

These studies demonstrate that narrative understanding often requires explicit representations of entities, events, temporal relations, causal structure, and changing world states. Their primary objective, however, is to support narrative interpretation, reasoning, or generation. The present work builds upon these representational foundations while asking how the information conveyed through a narrative can be reconstructed and maintained as a persistent computational world that remains available throughout subsequent interactive realization.

\subsection{Narrative-Conditioned Interactive Content Generation}

Narrative-conditioned content generation transforms narrative descriptions into interactive artifacts at multiple computational levels. Existing work generates scenes~\cite{chen2025narrative}, environments~\cite{kumaran2023scenecraft,nasir2024word2world,buongiorno2024pangea}, and complete interactive fiction games~\cite{zhou2025story2game}. Complementary research investigates narrative planning~\cite{ware2021sabre,siler2025pareto,fisher2024model,farrell2024large}, experience management~\cite{ware2022multiagent}, and alignment between gameplay progression and narrative goals~\cite{cardona2023aligning,rivera2024story}.

These approaches construct plans, scenes, environments, gameplay structures, or complete interactive experiences from narrative descriptions. In doing so, they often infer information not explicitly stated in the source narrative, such as spatial organization, object placement, event preconditions, affordances, or gameplay constraints. Such information is usually introduced to satisfy the requirements of a particular output or downstream task. This work instead makes the persistent world itself the primary computational object, allowing inferred and observed information to be maintained across multiple scenes, events, and interactions rather than reconstructed separately for each output.

\subsection{Procedural Content Generation and Mixed-Initiative Design}

Procedural Content Generation (PCG) develops computational methods for generating game content~\cite{summerville2018procedural,yannakakis2018artificial}. Recent research emphasizes intermediate representations throughout PCGML pipelines~\cite{guzdial2025pcgml}, mixed-initiative workflows~\cite{guzdial2025mixed}, semantic models of game systems~\cite{cardona2022game}, co-creative AI systems~\cite{lin2023ontology,agarwal2023controllable}, automated game design~\cite{cook2022puck,cook2025game}, and foundation-model-based PCG~\cite{gallotta2024large,yannakakis2025procedural}. These studies show that explicit intermediate representations can support controllability, author interaction, semantic organization, and coordination among different stages of game creation.

The persistent world considered in this work is related to such intermediate representations but serves a more specific role. It is not merely a temporary format used to connect stages of a generation pipeline. It maintains narrative-grounded entities, locations, relationships, and evolving states across narrative events and player interactions. It therefore provides a continuing semantic reference from which different interactive realizations may be produced and updated.





\subsection{Positioning Persistent World Reconstruction from Narratives}

Prior research provides computational representations for narrative understanding, planning, scene synthesis, environment construction, gameplay generation, and mixed-initiative design~\cite{gongora2026world,ware2021sabre,zhou2025story2game}. These representations capture many components also required by a persistent world, including entities, events, spatial relations, temporal structure, affordances, and evolving world states. The distinction lies not primarily in the information represented, but in the computational role that representation serves.

Existing approaches generally construct representations to support particular downstream tasks, such as interpreting a narrative, answering questions, generating a plan, synthesizing a scene, or producing an interactive experience. In contrast, persistent world reconstruction formulates the computational problem of constructing and maintaining the shared world implied by a narrative. The reconstructed persistent world preserves narrative observations, incorporates only the contextual information required for coherent interaction, maintains continuity as narrative observations and player interactions accumulate, and serves as the common computational foundation for subsequent interactive realization.

From this perspective, narrative understanding and interactive realization become complementary computational processes rather than independent stages. Narrative understanding provides observations and inferred structure from which the persistent world is reconstructed, while interactive realization instantiates and updates that maintained world. Persistent world reconstruction, therefore, defines the computational interface connecting these processes rather than introducing another task-specific intermediate representation.

%% file: 4_Framework.tex
\label{sec:framework}

\begin{figure*}[t]
    \centering
    \includegraphics[width=0.90\textwidth]{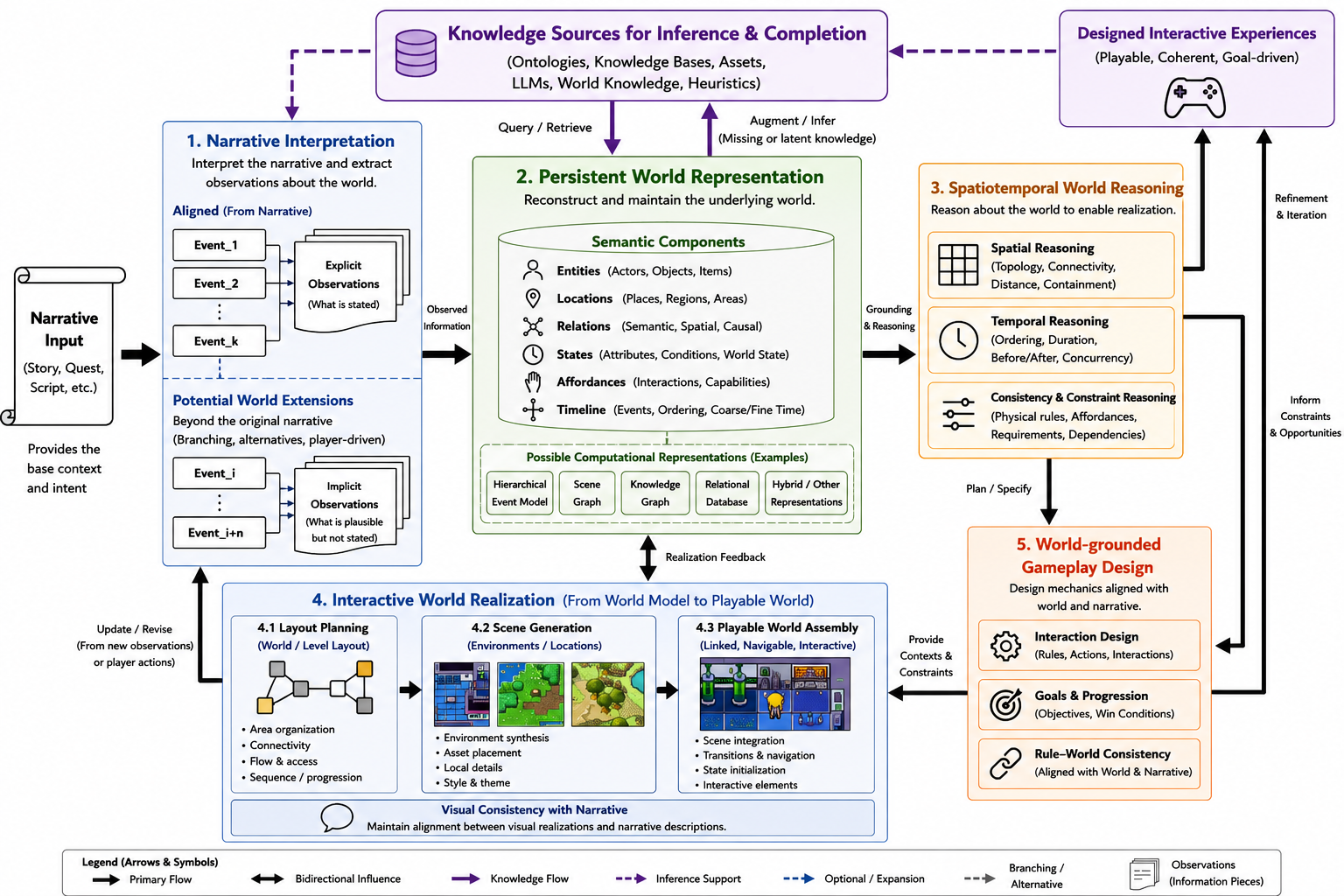}
    \caption{
    Reference framework for persistent world reconstruction from narrative descriptions. Narrative interpretation extracts structured observations, which are integrated through constrained world completion to reconstruct a persistent world. Spatiotemporal reasoning derives spatial organization and interaction constraints from the reconstructed world, supporting interactive realization and world-grounded gameplay. Player interactions and newly introduced observations subsequently update the same persistent world while preserving narrative consistency.
    }
    \label{fig:framework}
\end{figure*}

To operationalize the proposed formulation, we introduce a reference framework that organizes persistent world reconstruction from narrative descriptions into a sequence of computational stages. Rather than directly generating scenes, levels, or complete games from narrative input, the framework places an explicit persistent world at the center of computation, allowing subsequent reasoning and interactive realization to operate on a shared semantic representation. As illustrated in Figure~\ref{fig:framework}, the framework decomposes the process into narrative interpretation, persistent world construction, world reasoning, interactive realization, and world-grounded gameplay.

The framework specifies one conceptual organization of these computational stages rather than a fixed implementation pipeline. Individual stages may be implemented using different language models, symbolic reasoning systems, knowledge representations, or realization techniques without changing the underlying computational objective of reconstructing and maintaining a persistent world. The reference prototype described in the following section instantiates one implementation of this framework.



\subsection{Framework Overview}

The reference framework consists of five computational stages: (1) narrative interpretation and observation extraction, (2) persistent world construction, (3) spatiotemporal world reasoning, (4) interactive world realization, and (5) world-grounded gameplay design.

Narrative interpretation extracts observations describing entities, locations, actions, semantic relationships, and state changes from the source narrative. Persistent world construction integrates these observations into a maintained persistent world through constrained world completion, reconstructing only the contextual information required for coherent interaction while remaining grounded in the narrative evidence. Spatiotemporal world reasoning derives spatial organization, temporal consistency, and interaction constraints from the reconstructed world. Interactive realization instantiates portions of the maintained world as interactive experiences, while world-grounded gameplay derives interactions and gameplay mechanics from that same world. Subsequent player interactions and newly introduced narrative observations update the persistent world, allowing it to remain the shared computational state throughout continued interaction.

Collectively, these stages describe one conceptual organization of persistent world reconstruction rather than a fixed implementation pipeline. Different implementations may realize individual stages using different computational techniques while preserving the same underlying computational responsibilities.

\subsection{Narrative Interpretation and Observation Extraction}

Within the framework, the narrative interpretation stage treats a narrative as observations describing an evolving world rather than as a complete specification of an interactive environment. The stage extracts explicit entities, locations, actions, semantic relationships, and state changes and organizes them into structured observations. Because narratives intentionally omit information unnecessary for discourse, these observations provide only partial evidence about the underlying world. Additional observations introduced through branching interactions or author-directed modifications can subsequently be incorporated into the same persistent world.



\subsection{Persistent World Construction}

The persistent world construction stage integrates structured narrative observations into a maintained persistent world through constrained world completion. Rather than reconstructing every possible aspect of the underlying world, this stage reconstructs only the contextual information required to support coherent interactive realization while remaining grounded in the narrative evidence. The resulting persistent world maintains entities, locations, semantic relationships, evolving world states, and temporal organization as observations accumulate over time.

The framework specifies the computational responsibility of this stage rather than a particular reconstruction strategy. Knowledge graphs, symbolic world models, scene graphs, relational databases, hybrid symbolic-neural representations, or other structured representations are all compatible with the formulation, provided they support constructing and maintaining a persistent world that satisfies the properties defined in Section~\ref{subsec:persistent_world_reconstruction}. Different implementations may therefore realize this stage using different representations while preserving the same computational objective.

\subsection{Spatiotemporal World Reasoning}

The spatiotemporal reasoning stage derives the structural information required for interactive realization from the reconstructed persistent world. Spatial reasoning determines environmental layouts, navigation, containment relationships, and location connectivity, while temporal reasoning maintains event ordering, causal dependencies, and evolving world states. Consistency checks ensure that inferred knowledge remains compatible with both the reconstructed world and the original narrative. Because reasoning operates on the maintained persistent world rather than repeatedly interpreting the source narrative, information accumulated from earlier observations remains available throughout subsequent reasoning and interaction.



\subsection{Interactive World Realization}

The interactive realization stage instantiates portions of the reconstructed persistent world as interactive experiences. Rather than reconstructing world information independently for each realization, this stage operates on the maintained persistent world to derive the scene-level structures, interaction context, and state information required by a particular interactive experience.

Because realization is separated from persistent world construction, multiple layouts, visual presentations, interaction paradigms, or gameplay experiences may be instantiated from the same reconstructed world while remaining grounded in shared entities, locations, semantic relationships, and evolving world states. The framework, therefore, separates maintaining the world from realizing individual interactive experiences, allowing different realizations to reuse and subsequently update the same persistent world.

\subsection{World-Grounded Gameplay Design}

The world-grounded gameplay design stage derives gameplay mechanics from the reconstructed persistent world rather than authoring them independently. Interaction opportunities, objectives, progression constraints, and state-dependent behaviors are grounded in persistent entities, affordances, semantic relationships, and evolving world states. The persistent world specifies what exists and how it evolves, while gameplay specifies how players interact with that world. Consequently, the same persistent world may support multiple gameplay experiences while remaining grounded in the same narrative.

%% file: 5_Prototype.tex
\label{sec:prototype}

\begin{figure*}[t]
\centering

\begin{tabular}{ccc}

\textbf{Laboratory Entrance} &
\textbf{Workbench Preparation} &
\textbf{Mixing and Observation}
\\[0.3em]

\includegraphics[width=0.25\textwidth]{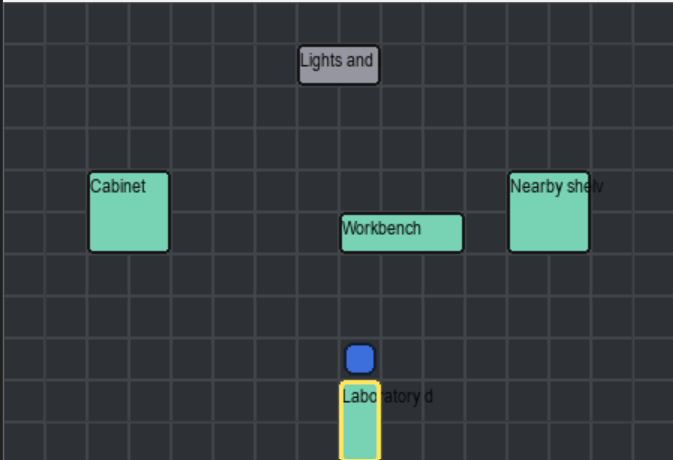} &
\includegraphics[width=0.25\textwidth]{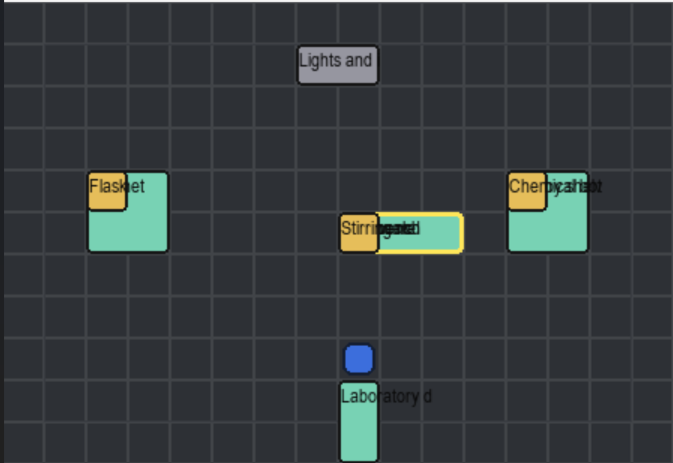} &
\includegraphics[width=0.25\textwidth]{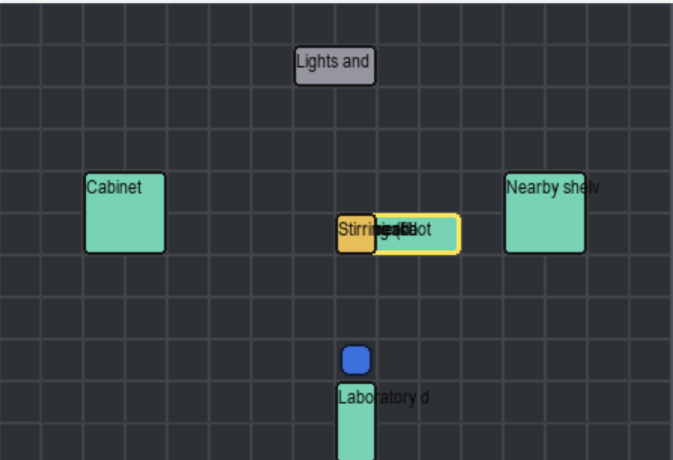}
\\

(a) Layout Realization &
(b) Layout Realization &
(c) Layout Realization
\\[0.8em]

\includegraphics[width=0.25\textwidth]{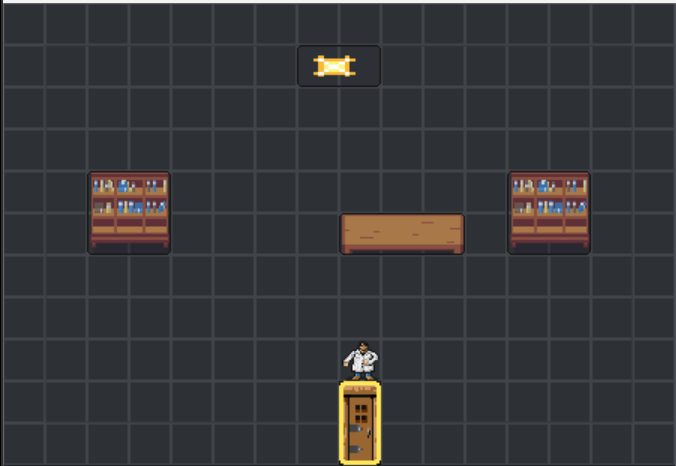} &
\includegraphics[width=0.25\textwidth]{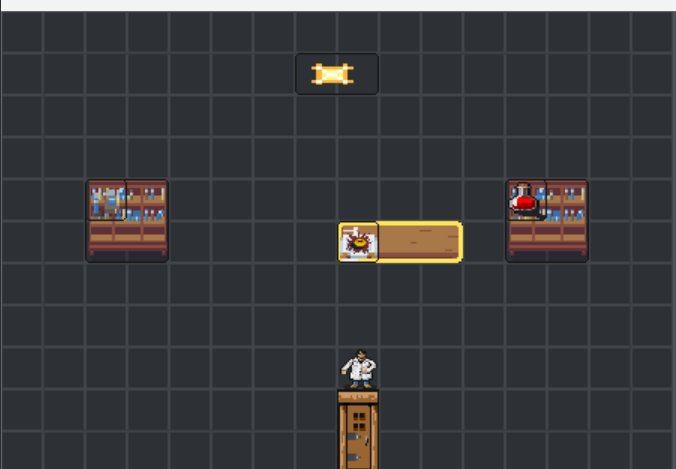} &
\includegraphics[width=0.25\textwidth]{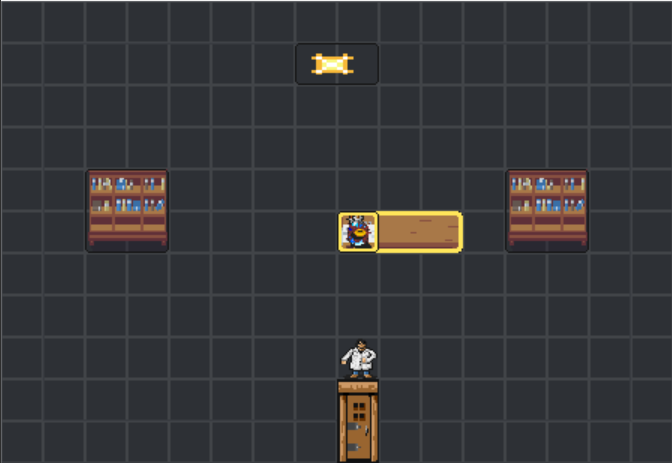}
\\

(d) Asset-Grounded Scene &
(e) Asset-Grounded Scene &
(f) Asset-Grounded Scene

\end{tabular}

\caption{
Realization results for the \emph{Chemical Laboratory} prototype. The top row visualizes tile-based layouts derived from the reconstructed persistent world, and the bottom row shows the corresponding asset-grounded scenes. Persistent entities and evolving world states remain synchronized across the three scenes.
}
\label{fig:gameplay_sequence}

\end{figure*}

\begin{table}[t]
\centering

\begin{tabular}{p{0.20\columnwidth}p{0.2\columnwidth}p{0.45\columnwidth}}
\textbf{Case} &
\textbf{Evaluation Scope} &
\textbf{Demonstrated Capability} \\
\hline

Chemical Laboratory &
End-to-end realization &
Persistent world reconstruction, scene realization, and world-grounded gameplay. \\
\hline

The Forgotten Shrine &
Multi-location persistence &
Persistence across connected locations, scene transitions, and navigation. \\
\hline

Little Red Riding Hood &
Existing narrative adaptation &
Applicability to a narrative adapted from an existing story. \\

\end{tabular}
\caption{Prototype cases used to examine complementary aspects of persistent world reconstruction and interactive realization.}
\label{tab:prototype_cases}
\end{table}

The reference framework presented in Section~\ref{sec:framework} identifies the computational stages required to reconstruct and use a persistent world for narrative-grounded interactive experiences. This section presents a reference prototype implementing that framework. Rather than serving as a production-ready game-generation system, the prototype provides one operational realization of the proposed formulation, demonstrating how an explicit persistent world can be reconstructed from narrative observations and subsequently support scene realization, gameplay, and interaction updates.

The prototype is designed to examine three aspects of the proposed formulation:

\begin{enumerate}
    \item reconstructing and maintaining a persistent world from narrative observations;
    \item instantiating portions of the reconstructed world as playable interactive environments while preserving continuity across events, scenes, and locations; and
    \item grounding gameplay mechanics and interaction updates in persistent entities, locations, affordances, semantic relationships, and evolving world states.
\end{enumerate}

The implementation intentionally simplifies several realization components so that the evaluation focuses on the computational role of the persistent world rather than the performance of individual generation modules. Narrative interpretation builds upon structured information extraction~\cite{chen2025narrative}, semantic asset grounding employs GameTileNet~\cite{chen2025gametilenet}, and spatial layout realization, together with gameplay mechanics, uses lightweight rule-based procedures. The remainder of this section introduces the prototype cases and describes how the reference framework is instantiated in the current implementation.




\subsection{Prototype Cases}
\label{subsec:prototype_cases}

The reference prototype is demonstrated through three narrative cases chosen to exercise complementary aspects of persistent world reconstruction and interactive realization rather than to serve as benchmark datasets. Each case begins with a manually designed event specification that is realized as a natural-language narrative using \textit{OpenAI GPT-5-mini} and a shared prompt template. The prompt template and generated narratives are provided in the Appendix.

The three cases emphasize different capabilities of the prototype implementation. \emph{Chemical Laboratory} serves as the primary running example because it exercises the complete pipeline from narrative interpretation and persistent world reconstruction to interactive realization. \emph{The Forgotten Shrine} focuses on persistence across multiple connected locations, including inventory continuity and world-state maintenance during scene transitions. \emph{Little Red Riding Hood} demonstrates that the same implementation can be applied to a narrative adapted from an existing public-domain story rather than a procedurally authored scenario.

Together, these cases demonstrate how the proposed formulation can be instantiated through the reference prototype while exercising different aspects of persistent world reconstruction. Table~\ref{tab:prototype_cases} summarizes the three cases, while additional intermediate representations and realization results are included in the Appendix.

\subsection{Narrative Interpretation}
\label{subsec:narrative_interpretation}

The first stage transforms a narrative into structured observations that provide evidence for persistent world construction. It uses a hierarchical representation adapted from prior work on visual narrative understanding~\cite{chen2025hierarchical}. Because the evaluation narratives are relatively short, the prototype retains only the event and event-fragment levels. As summarized in Table~\ref{tab:narrative_hierarchy}, events correspond to candidate playable situations organized around immediate objectives, while event fragments capture the actions and observations contributing to those objectives.

Narrative interpretation is implemented using a prompt-based large language model following the structured extraction strategy of Narrative-to-Scene Generation~\cite{chen2025narrative}. Under a constrained JSON schema, the model extracts the event hierarchy together with entities, locations, semantic relationships, object states, and temporal ordering explicitly supported by the narrative. No additional world knowledge is inferred during this stage. All extracted observations are manually verified before they are incorporated into the persistent world. Tables~\ref{tab:narrative_hierarchy} and~\ref{tab:narrative_interpretation_results} summarize the adopted representation and representative interpretation results.

\begin{table*}[t]
\centering

\begin{tabular}{p{0.18\textwidth}p{0.58\textwidth}p{0.18\textwidth}}
\textbf{Representation Level} &
\textbf{Description} &
\textbf{Role in the Prototype} \\
\hline

Narrative &
Complete natural-language story provided as input. &
Narrative input. \\
\hline

Macro-event &
High-level grouping of related events in longer narratives. &
Omitted in the current prototype. \\
\hline

Event &
Coherent narrative situation organized around one immediate objective. &
Candidate playable scene. \\
\hline

Event Fragment &
Individual actions or observations contributing toward the event objective. &
Scene interactions and state changes. \\
\hline

Narrative Observation &
Explicit entities, relationships, actions, and states extracted from event fragments. &
Evidence for world construction. \\

\end{tabular}
\caption{Hierarchical narrative representation adopted by the reference prototype.}
\label{tab:narrative_hierarchy}
\end{table*}

\begin{table*}[t]
\centering

\begin{tabular}{p{0.15\textwidth}p{0.15\textwidth}p{0.24\textwidth}p{0.35\textwidth}}
\textbf{Case} &
\textbf{Event} &
\textbf{Immediate Objective} &
\textbf{Representative Event Fragments} \\
\hline

Chemical Laboratory &
Prepare the Experiment &
Gather materials and prepare the workspace. &
Clear workbench; arrange tools; find flask; collect chemical bottles. \\
\hline

Chemical Laboratory &
Mix and Observe Reaction &
Mix chemicals and monitor the reaction. &
Pour and stir chemicals; observe color change; observe bubbling. \\
\hline

Forgotten Shrine &
Find the Bronze Key &
Search the watchtower and obtain the key. &
Travel to watchtower; search rooms; find key in chest; keep key. \\
\hline

Forgotten Shrine &
Enter the Shrine &
Use the key to access the shrine and retrieve the crystal. &
Reach shrine; clear vines; unlock entrance; retrieve crystal. \\
\hline

Little Red Riding Hood &
Encounter the Wolf &
Navigate the encounter and continue toward the cottage. &
Meet wolf; observe encounter; continue to cottage. \\
\hline

Little Red Riding Hood &
Rescue Grandmother &
Find grandmother in danger and make the cottage safe. &
Open cottage door; discover danger; move bedcover; help grandmother; secure room. \\

\end{tabular}
\caption{Representative narrative interpretation results. Actions and observations contributing to the same immediate objective are grouped into a single event, providing the basis for persistent world construction and subsequent scene realization.}
\label{tab:narrative_interpretation_results}
\end{table*}

\subsection{Persistent World Construction}
\label{subsec:persistent_world_construction}

Narrative observations describe actions and local state changes but do not by themselves constitute a persistent world. The prototype therefore normalizes narrative observations into symbolic world facts and incrementally integrates those facts into an evolving world representation.

State normalization converts narrative actions into persistent facts whenever possible. For example, \emph{Scientist entered the laboratory} becomes the spatial relation \emph{Scientist inside Laboratory}, while \emph{Scientist opened the cabinet} changes the cabinet state to \emph{Open}. Representative examples are shown in Table~\ref{tab:state_normalization}.

The normalized facts are accumulated into three complementary structures: a hierarchical event representation preserving narrative and temporal organization~\cite{chen2025hierarchical}, a location graph describing navigable locations and their semantic connections, and an entity-state representation maintaining persistent identities, locations, ownership, properties, and evolving states. Together, these structures implement the persistent world used by the prototype. As observations are incorporated, recurring entities retain their identities and previously established states unless they are explicitly modified by later narrative observations or player interactions.

\begin{table}[t]
\centering

\begin{tabular}{p{0.33\columnwidth}p{0.3\columnwidth}p{0.22\columnwidth}}
\textbf{Narrative Observation} &
\textbf{Normalized World Fact} &
\textbf{Purpose} \\
\hline

Scientist entered the laboratory &
Scientist \texttt{inside} Laboratory &
Persistent spatial state \\
\hline

Scientist opened the cabinet &
Cabinet \texttt{state = open} &
Persistent object state \\
\hline

Scientist picked up the flask &
Scientist \texttt{holding} Flask &
Track movable entities \\
\hline

Scientist placed the flask on the workbench &
Flask \texttt{on} Workbench &
Maintain spatial continuity \\

\end{tabular}

\caption{Representative examples of state normalization. Narrative observations are transformed into persistent world facts that maintain entity identity, spatial relationships, and evolving states across events and interactions.}
\label{tab:state_normalization}

\end{table}

\subsection{World Completion}
\label{subsec:world_completion}

Narratives omit many details required for interactive realization, including environmental structures, spatial context, object affordances, and interaction constraints. The persistent world must therefore include not only explicitly communicated observations but also the contextual structures needed to make those observations interactively realizable. World completion supplies this missing context while remaining grounded in both the narrative evidence and the world facts already reconstructed.

The prototype performs three forms of constrained inference: (1) interaction-required world knowledge, such as a workbench for chemical mixing or a doorway connecting locations; (2) placement constraints distinguishing fixed and movable entities; and (3) realization hints describing generic environmental characteristics. Representative examples are summarized in Table~\ref{tab:world_completion}.

Candidate completions are produced using a prompt-based large language model conditioned on the reconstructed persistent world. Rather than producing unrestricted story continuations, the model proposes structured entities, relationships, affordances, environmental context, and placement constraints. All proposed additions are manually verified before they are incorporated into the persistent world.

Before interactive realization, the completed persistent world is projected into scene-level representations. Each projection selects the entities, interaction targets, objectives, and placement constraints required for a particular playable scene. Persistent entities and their evolving states are propagated across scene projections unless they have been modified by later narrative observations or interactions. This projection preserves global continuity while exposing only the portion of the persistent world required by each scene.

\begin{table}[t]
\centering

\begin{tabular}{p{0.18\columnwidth}p{0.24\columnwidth}p{0.42\columnwidth}}
\textbf{Inference} &
\textbf{Example} &
\textbf{Role} \\
\hline

Interaction knowledge &
Workbench for chemical mixing &
Support an interaction required by the narrative. \\
\hline

Placement constraint &
Cabinet = fixed; bottles = movable &
Guide layout without prescribing exact coordinates. \\
\hline

Realization hint &
Laboratory floor; shrine; forest &
Guide visual and environmental realization. \\

\end{tabular}

\caption{Representative forms of constrained world completion used by the prototype.}
\label{tab:world_completion}

\end{table}

\begin{table}[t]
\centering

\begin{tabular}{p{0.15\columnwidth}p{0.25\columnwidth}p{0.22\columnwidth}p{0.21\columnwidth}}
\textbf{Mechanic} &
\textbf{Required World Information} &
\textbf{Example} &
\textbf{Updated State} \\
\hline

Inspect &
Entity attributes &
Inspect cabinet &
None \\
\hline

Collect &
Entity location and ownership &
Collect flask &
Inventory updated \\
\hline

Use &
Affordances and object relations &
Mix chemicals &
Entity state updated \\
\hline

Transition &
Task completion and location graph &
Enter next location &
Player location updated \\

\end{tabular}

\caption{Gameplay mechanics implemented in the prototype and the persistent-world information required to support each interaction.}
\label{tab:mechanics}
\end{table}

\subsection{Visual and Interactive Realization}
\label{subsec:visual_interactive_realization}

The scene-level projections produced from the persistent world are instantiated as playable tile-based environments. Unlike the persistent world, which maintains global semantic and state information throughout the narrative, each scene representation contains only the entities, interaction targets, placement constraints, and gameplay objectives required for a particular playable situation. This separation decouples persistent world maintenance from engine-specific realization while preserving continuity across scenes.

Spatial realization converts each scene representation into a navigable tile-based layout. Fixed entities are positioned according to inferred placement constraints, while movable entities are initialized from the persistent states propagated during scene projection. Because different scenes refer to the same persistent world, they preserve shared entity identities while reflecting changes in location, ownership, inventory, and object state.

Visual realization is performed through semantic asset grounding with GameTileNet~\cite{chen2025gametilenet}. Scene entities are matched to candidate tile assets using semantic labels and affordances. The prototype retrieves the top-$k$ candidates and applies lightweight manual verification before selecting the final asset for each entity.

The realized scenes are assembled into a playable PyGame prototype. Rather than generating gameplay automatically, the prototype implements a small set of reusable mechanics that query and update the reconstructed persistent world. The current implementation supports object inspection, item collection, inventory management, object interaction, object use and combination, and location transitions. These interactions update persistent entity states, locations, and inventory contents, allowing later scenes to remain synchronized through the shared world rather than through independent scene-specific scripts. Figure~\ref{fig:gameplay_sequence} presents representative realization results, while Table~\ref{tab:mechanics} summarizes the persistent-world information required by each implemented mechanic.



\subsection{Artifact and Reproducibility}
\label{subsec:artifact}

A public repository accompanies this manuscript. \footnote{\textbf{Repo:} \url{https://github.com/RimiChen/2026_Narrative2World}}  The repository contains the reference prototype implementation together with the narrative cases, prompt templates, intermediate representations, and scripts required to reproduce the reported demonstrations.

The current implementation uses \textit{OpenAI GPT-5-mini} to realize manually designed event specifications as natural-language narratives, extract structured narrative observations, and propose constrained world completions. All model-generated artifacts incorporated into the reported prototype are manually verified before being integrated into the persistent world. Appendix additionally provide complete intermediate representations and realization results for all three prototype cases.

%

%% file: 6_Evaluation.tex
\label{sec:evaluation}

The proposed formulation is examined through the reference prototype presented in Section~\ref{sec:prototype}. Rather than benchmarking a task-specific generation algorithm, the validation examines whether a reference implementation of the formulation can reconstruct an explicit persistent world from narrative observations and use that world to support coherent interactive realization. The three prototype cases exercise complementary aspects of the formulation, including end-to-end reconstruction, persistence across multiple connected locations, and applicability beyond the procedural scenarios used during prototype development.

The \emph{Chemical Laboratory} case serves as the primary end-to-end demonstration of the reference prototype. Figure~\ref{fig:gameplay_sequence} presents representative realization results. Narrative observations are reconstructed into a persistent world, from which scene-level representations are projected and instantiated as playable tile-based environments. Throughout the three scenes, entity identities, spatial relationships, inventory contents, and object states remain synchronized through the shared persistent world. Player interactions subsequently update that same world, allowing later scenes to reflect changes introduced during earlier gameplay.

The \emph{Forgotten Shrine} case examines persistence across multiple connected locations. Rather than emphasizing gameplay complexity, this case demonstrates that persistent entities, inventory contents, player location, and object states remain consistent across navigation and scene transitions through the maintained persistent world. For example, once the bronze key is collected, it remains associated with the player and is subsequently available when interacting with the shrine. Representative intermediate representations and realization results are included in the Appendix.

The \emph{Little Red Riding Hood} case examines whether the same prototype implementation can be applied to narratives adapted from existing stories. Using the same interpretation, reconstruction, world completion, and realization pipeline, the narrative is transformed into a persistent world without modifying the underlying framework. Although this case does not establish broad generality across narrative genres, it demonstrates that the proposed formulation is not restricted to procedurally authored scenarios.

Together, the three case studies demonstrate that the reference prototype can operationalize the proposed formulation by reconstructing an explicit persistent world that serves as a shared computational foundation for narrative interpretation, constrained world completion, scene realization, gameplay mechanics, and interaction updates. The validation is intentionally qualitative. Rather than evaluating visual quality, gameplay sophistication, or generation diversity, it examines whether information reconstructed from narrative observations remains available, coherent, and reusable throughout interactive realization. The objective is therefore to assess the feasibility of the proposed formulation through its reference implementation rather than to optimize or benchmark individual realization components.

%% file: 7_Discussion.tex
\section{Discussion}

The reference prototype was developed to investigate the computational role of persistent worlds in narrative-grounded interactive experiences rather than to optimize individual realization components. Although intentionally lightweight, implementing the prototype provided several insights into reconstructing and maintaining persistent worlds for interactive realization.

First, the prototype intentionally adopts a minimal interactive realization rather than a visually sophisticated game engine. The objective of this work is not to evaluate graphical fidelity or gameplay complexity, but to examine whether an explicit persistent world can coordinate narrative interpretation, scene realization, gameplay interactions, and evolving world states. A lightweight tile-based environment provides sufficient expressive power to exercise persistence, state propagation, interaction updates, and cross-scene consistency while minimizing implementation complexity unrelated to the proposed formulation.

Second, continuity across narrative progression proved to be more fundamental than reconstructing individual observations in isolation. Early prototype versions realized narrative events independently, causing entities to disappear between consecutive scenes despite remaining part of the same underlying world. The final prototype, therefore, propagates persistent entities and evolving world states during scene projection, allowing information established by earlier narrative observations to remain available until explicitly modified by later narrative events or player interactions. This experience suggests that temporal persistence is a fundamental computational requirement rather than simply a representation choice.

Third, the implementation highlights the importance of separating the persistent world from its interactive realizations. The persistent world maintains global information describing entities, locations, semantic relationships, affordances, and evolving world states, whereas each scene representation contains only the subset of information required for a particular interactive situation. This separation allows multiple scenes, layouts, interaction structures, or gameplay experiences to be instantiated from the same reconstructed world while preserving overall narrative consistency.

Fourth, implementing constrained world completion revealed that interactive realization requires contextual information extending beyond explicit narrative observations. Environmental structures, placement constraints, affordances, and navigational context are often necessary to support coherent interaction even though they are not directly described by the narrative. Conditioning world completion on the reconstructed persistent world helps ensure that inferred information remains compatible with previously established entities, relationships, and world states.

More broadly, the reference implementation suggests that the persistent world should be regarded as a long-lived computational object rather than merely an intermediate representation generated for a particular downstream task. Narrative interpretation reconstructs this world from observations, interactive realization consumes it to instantiate scenes and gameplay, and subsequent player interactions continue to update the same world. The persistent world, therefore, functions as the shared computational state connecting narrative understanding with interactive realization throughout the lifetime of an experience.

The proposed formulation also opens several directions for future research. At the implementation level, richer world representations, more sophisticated reasoning mechanisms, and support for long-running interactive narratives would substantially extend the capabilities of the current prototype. More broadly, because the formulation is independent of any particular implementation strategy, future work may investigate integrating explicit persistent worlds with end-to-end foundation models. Rather than replacing explicit world reconstruction, such architectures could use persistent worlds as stable computational references that maintain entities, locations, semantic relationships, and evolving world states throughout the interpretation and interaction of extended narratives.

\section{Limitations}

The proposed formulation is examined through a lightweight reference prototype, and several limitations remain.

First, the current implementation intentionally simplifies several realization components, including rule-based gameplay mechanics, lightweight spatial layout generation, and manual verification during semantic asset grounding. These choices isolate the proposed formulation from implementation-specific complexity, but they limit the visual richness, gameplay sophistication, and degree of automation of the resulting experiences.

Second, narrative interpretation and constrained world completion currently rely on prompt-based large language models together with manual verification. The reported demonstrations, therefore, establish the feasibility of the proposed formulation rather than fully automatic persistent world reconstruction from unrestricted narrative text. More robust information extraction, commonsense reasoning, provenance tracking, and automatic consistency verification remain important engineering challenges.

Third, the current validation consists of three representative case studies chosen to exercise complementary aspects of the formulation. Although these demonstrate proof-of-concept feasibility, broader evaluations involving longer narratives, more diverse interactive environments, and richer gameplay experiences will be necessary to characterize scalability and generality.

Finally, while this work demonstrates that a maintained persistent world can support coherent interactive realization, it does not quantitatively compare this approach against alternative generation strategies that reconstruct scenes independently. Future evaluations comparing continuity, contradiction rates, state consistency, and downstream interaction quality would provide stronger evidence for the computational benefits of explicit persistent world reconstruction.

%% file: 8_Conclusion.tex
This paper formulated \emph{persistent world reconstruction from narrative descriptions} as a computational problem for supporting narrative-grounded interactive experiences. Rather than treating narratives as direct specifications of scenes, levels, or complete games, the proposed formulation reconstructs and maintains an explicit persistent world that serves as the shared computational object connecting narrative understanding with interactive realization.

To investigate this formulation, we presented a reference framework and implemented a reference prototype that reconstructs persistent worlds from narrative observations before instantiating playable tile-based environments. Through three complementary case studies, the prototype demonstrates the feasibility of reconstructing and maintaining persistent entities, locations, semantic relationships, and evolving world states while supporting coherent interactive realization.

More broadly, this work argues that persistent world reconstruction should be regarded as a distinct computational problem rather than an implementation detail embedded within individual generation tasks. The reconstructed persistent world provides the long-lived computational foundation that grounds narrative observations, contextual completion, interactive realization, and subsequent interaction updates in the same evolving world. We hope this formulation provides a foundation for future research into richer persistent-world representations, scalable reconstruction methods, long-running interactive narratives, and the integration of explicit persistent worlds with future AI systems.